\documentclass[letterpaper, 10pt, journal, twoside]{./Style_template/RSS/IEEETrans}  

\usepackage[letterpaper, left=0.7in, right=0.7in, bottom=0.76in, top=0.7in] {geometry}
\def\IEEEtitletopspace{0.2in}
\IEEEoverridecommandlockouts                              
\usepackage{cite}

\usepackage{amsmath}
\usepackage{moreverb,url}
\usepackage{times}
\usepackage{multicol}
\usepackage{graphicx}
\usepackage{amssymb} 
\usepackage{gensymb}
\usepackage{amsmath, xparse}
\usepackage{mathrsfs}
\usepackage{soul}
\usepackage{enumitem,kantlipsum}
\usepackage{multirow}
\usepackage[colorlinks,bookmarksopen,bookmarksnumbered,citecolor=red,urlcolor=blue]{hyperref}

\newcommand{\Revision}{\color[rgb]{0, 0, 0}} 

\newcommand\BibTeX{{\rmfamily B\kern-.05em \textsc{i\kern-.025em b}\kern-.08em
T\kern-.1667em\lower.7ex\hbox{E}\kern-.125emX}}

\def\ie{{\em i.e.,\ }}

\begin{document}

\title{A bio-inspired sand-rolling robot: effect of body shape on sand rolling performance}

\author{Xingjue Liao$^{1}$, Wenhao Liu$^{1}$, Hao Wu$^{1}$ and Feifei Qian$^{*,1}$
\thanks{$^{1}$Xingjue Liao, Wenhao Liu, Hao Wu and Feifei Qian are with Department of Electrical and Computer Engineering, University of Southern California, Los Angeles, CA, USA.
        {\tt\footnotesize feifeiqi@usc.edu}
        (\textit{corresponding author: Feifei Qian})}%
\thanks{This research is supported by funding from the National Science Foundation (NSF) CAREER award \#2240075, the NASA Planetary Science and Technology Through Analog Research (PSTAR) program, Award \# 80NSSC22K1313, and the NASA Lunar Surface Technology Research (LuSTR) program, Award \# 80NSSC24K0127.}

}

\maketitle

\begin{abstract}

The capability of effectively moving on complex terrains such as sand and gravel can empower our robots to robustly operate in outdoor environments, and assist with critical tasks such as environment monitoring, search-and-rescue, and supply delivery. Inspired by the Mount Lyell salamander's ability to curl its body into a loop and effectively roll down {\Revision hill slopes}, in this study we develop a sand-rolling robot and investigate how its locomotion performance is governed by the shape of its body. We experimentally tested three different body shapes: Hexagon, Quadrilateral, and Triangle. We found that Hexagon and Triangle can achieve a faster rolling speed on sand, but exhibited more frequent failures of getting stuck. Analysis of the interaction between robot and sand revealed the failure mechanism: the deformation of the sand produced a local ``sand incline'' underneath robot contact segments, increasing the effective region of supporting polygon (ERSP) and preventing the robot from shifting its center of mass (CoM) outside the ERSP to produce sustainable rolling. Based on this mechanism, a highly-simplified model successfully captured the critical body pitch for each rolling shape to produce sustained rolling on sand, and informed design adaptations that mitigated the locomotion failures and improved robot speed by more than 200$\%$.  Our results provide insights into how locomotors can utilize different morphological features to achieve robust rolling motion across deformable substrates.

\end{abstract}
\begin{IEEEkeywords}  
Biologically-Inspired Robots; Locomotion on Deformable Terrains; Contact Modeling
\end{IEEEkeywords}

\section{Introduction}\label{sec:Intro} 

\begin{figure}[htbp!]
\centering
\includegraphics[scale=0.45]{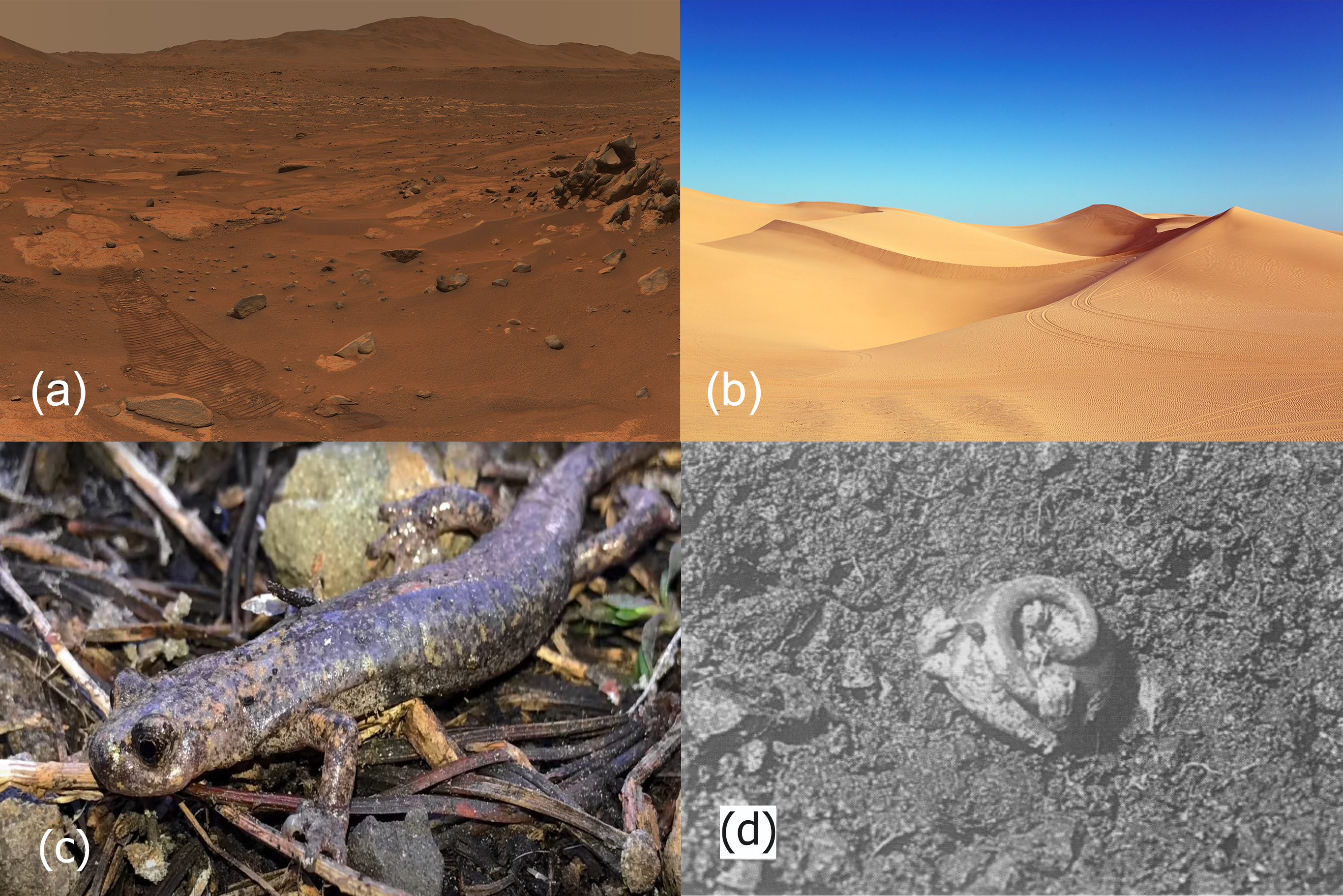}
\caption{Deformable substrates in Earth and extraterrestrial environments, and animals that exhibit high locomotion performance in these environments. (a) Loose regolith on Mars. Image credit NASA/JPL-Caltech/ASU/MSSS. (b) Sand dunes in California desert. Image credit Wikimedia Commons.  {\Revision (c, d) \textit{Hydromantes platycephalus} (Mount Lyell Salamander), in resting posture (c) and curled up rolling posture (d). Images reproduced from~\cite{garcia1995novel}.} 
}
\label{Fig.Introduction}
\end{figure}

Deformable substrates such as sand, dirt, and soil exist widely in both Earth and extraterrestrial environments (Fig. \ref{Fig.Introduction}a, b). The ability for robots to successfully and robustly move across deformable substrates can empower our robots for a variety of critical missions in agriculture, transportation, post-disaster search-and-rescue, and environment monitoring. However, moving on deformable substrates can be extremely challenging, as the substrates can exhibit both solid-like and fluid-like behaviors~\cite{jaeger1992physics}, leading to sinkage, slippage, and other complex locomotor-terrain interactions~\cite{bekker1956theory,wong1989terramechanics}. Misapplied locomotion strategies can lead to catastrophic failures~\cite{sorice2021insight}.


Many biological locomotors (Fig. \ref{Fig.Introduction}c, d) exhibit high locomotion performance on deformable substrates~\cite{maladen2009undulatory,mazouchova2013flipper,marvi2014sidewinding,qian2015principles,mcinroe2016tail,astley2020surprising}. Their morphology and strategies have inspired designs and controls for robot locomotion on complex terrains~\cite{maladen2011undulatory,li2009sensitive,qian2013walking,qian2015principles,marvi2014sidewinding,mcinroe2016tail,wu2019tactile,Haomachai2021,Guetta2023}. Among the high-performance animals on {\Revision loose substrates, the 
\textit{Hydromantes platycephalus} (Mount Lyell Salamander) ~\cite{downerweird, garcia1995novel} (Fig. \ref{Fig.Introduction}c, d)} 
uses an distinct strategy: by curling up their body into a circle, they are able to agily and stably roll down {\Revision muddy slopes} or rugged rocky hills.

Intrigued by the salamander's interesting locomotion mode and high performance, we seek to understand the principles governing self-propelled rolling performance on sand. Many previous studies have proposed designs for rolling robots. One common design is to control the joint angles between closed-chain segments to achieve rolling~\cite{SMA}. Another study proposed a mechanism that uses a center-arranged actuator to drive the connection between inner crank links and outer chain. Rather interestingly, the JSEL~\cite{JSEL} robot achieved self-propelled rolling by locally deforming its surface cells through jamming and unjamming. In most of these studies, the rolling robot was developed for and tested on relatively simple terrains that are mostly rigid and flat. How deformable substrates would influence rolling robot performance remains an under-explored question. 

To address this gap, in this paper we compared the locomotion performance of a rolling robot on rigid and deformable substrates (Sec. \ref{sec:Design}), and identified unique challenges for sand-rolling. To understand how sand-rolling performance was affected by robot morphology, we investigated three different robot shape kinematics. Based on observed robot-substrate interactions and resulting performance, we proposed a simplified model that revealed the failure mechanism of sand-rolling (Sec. \ref{sec:results}), and demonstrated model-informed design adaptations that resulted in significantly improved sand-rolling performance (Sec. \ref{sec:strategy}). {\Revision Lastly, we expanded our model to generalized robot shapes (Sec. \ref{sec:General Shape}). }

\section{Materials and Methods}\label{sec:Design}
\subsection{Robot} 

\begin{figure}[htbp]
\centering
\includegraphics[scale=0.6]{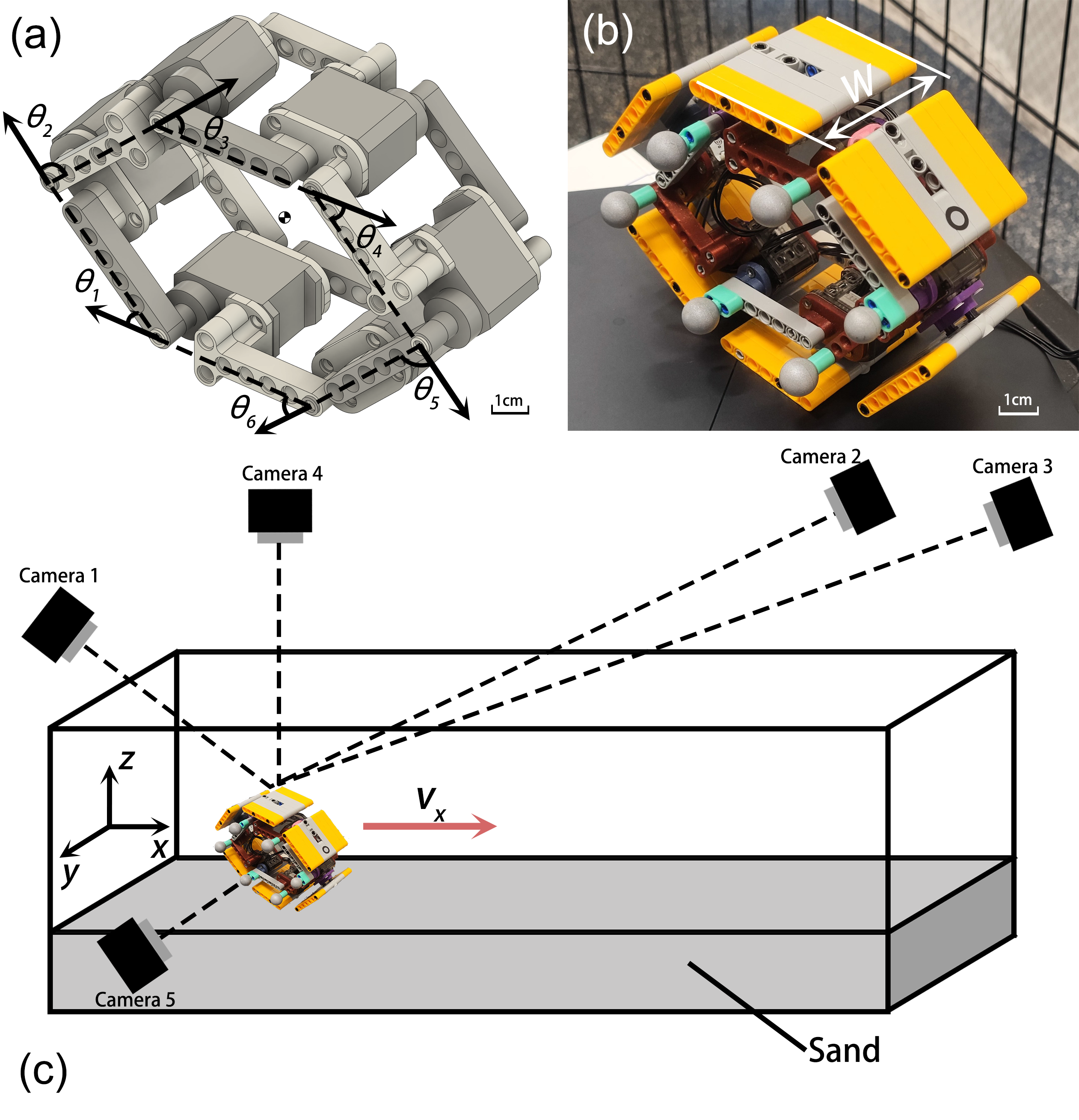}
\caption{Sand-rolling robot, and experimental setup to study its sand-rolling performance. (a) Simplified 3D model of the rolling robot. (b) Snapshot of the rolling robot. 6 reflective markers were attached at the six joints of the robot for tracking.  
(c) Diagram of the granular trackway for testing robot locomotion on sand. Camera 1-3 were used for record tracking data, whereas camera 4-5 were used to record top view and side view experimental footage.}
\label{Fig.structure}
\end{figure}

\begin{figure}[htbp]
\centering
\includegraphics[scale=0.3]{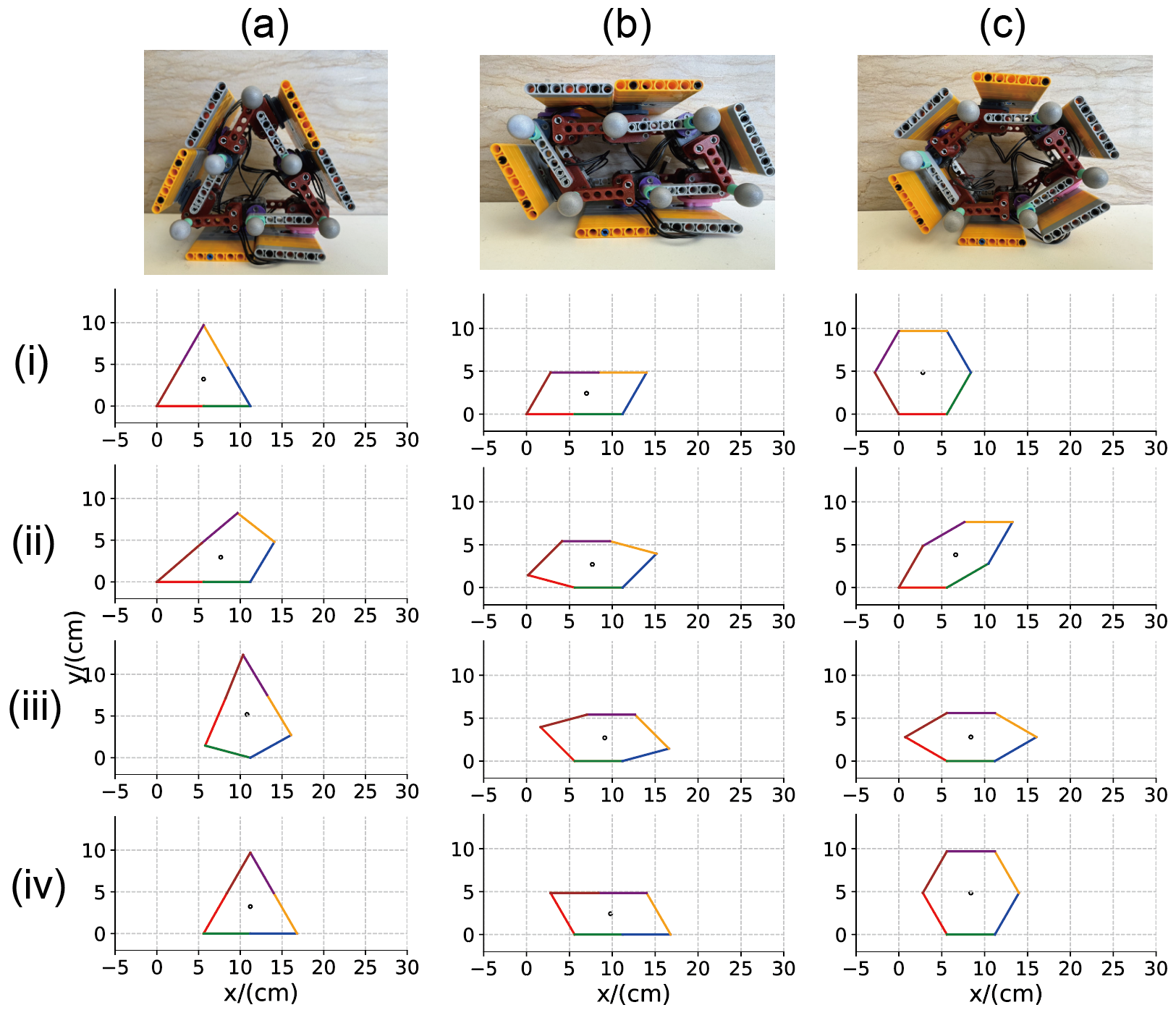}  
\caption{The kinematics of the three body shapes: (a) Triangle, (b) Quadrilateral, (c) Hexagon. 
}
\label{Fig.kinematics}
\end{figure}

To investigate how body morphology affects sand-rolling performance, we developed a bio-inspired robot as a robophysical~\cite{aguilar2016robophysical} model to perform controlled locomotion experiments. Modelled after the tumbling lizards which can bend their bodies into a closed loop, the sand-rolling robot was designed with 6 joints to form a closed chain (Fig. \ref{Fig.structure}a, b). Each joint was actuated by a servo motor (Dynamixel XL-320) which was controlled using a microcontroller (Arduino Uno). The length between each pair of adjacent joint axes was 5.6 cm. In this study, we were primarily interested in studying the rolling performance along the fore-aft direction, and thus the width of each segment was chosen to be sufficient wide ($W$ = 7.2 cm) to prevent rotation {\Revision about} the $x$-{\Revision axis}. Three robot body shapes were tested to study how body shape affects sand-rolling performance: a ``Triangle'' shape (Fig. \ref{Fig.kinematics}a), a ``Quadrilateral'' shape (Fig. \ref{Fig.kinematics}b), and a ``Hexagon'' shape (Fig. \ref{Fig.kinematics}c). 
The kinematics of the robot for each shape are shown in Fig.   \ref{Fig.kinematics}i-iv. A total of 36 trials were performed, 12 for each shape. Among those 12 trials, 6 were performed on rigid ground and another 6 on deformable substrates (see Sec. \ref{sec:setup}). For all trials in this study, the robot stride period was set to be 3 seconds. 

\subsection{Experiment Setup} \label{sec:setup}

Robot locomotion experiments were performed in a 125 cm long $\times$ 65 cm wide $\times$ 17 cm deep trackway (Fig. \ref{Fig.structure}c). Plastic ``sand particles'' {\Revision (Matrix Tactical Systems, Airsoft BB)} with 6 mm particle diameter were used as a model granular substrate in this study. The 6 mm plastic particles have been proven to be a promising model substrate for studying the interaction between robots and granular materials~\cite{maladen2009undulatory,maladen2011undulatory}, as they behave qualitatively similar in terms of rheology and forces~\cite{li2013terradynamics} with natural sand/soil with more angularity and size polydispersity, while the simpler geometry facilitates the control of the substrate to achieve uniform packing state~\cite{dickinson1994low}. To ensure a consistent initial condition for each locomotion test, prior to each trial we manually loosen the particles across the trackway prior to each trial, then flatten the surface to eliminate surface height variations. 

To characterize the robot speed during sand rolling, three motion capture cameras (Optitrack, Prime 13W) were installed around the trackway to record the position of each joint in the world frame ($x_i$ - fore-aft; $y_i$ - lateral; $z_i$ - vertical, where $i$ is the joint index) 
at 120 frames per second (FPS). Robot center-of-mass (CoM) position in the world frame ($x_c$ - fore-aft; $y_c$ - lateral; $z_c$ - vertical) was then computed as the geometric center of all joints. In addition, two 
video camera (Optitrack, Prime Color) were used to record experiment footage from both top view and side view at 120 FPS.

\section{Results and discussion}\label{sec:results} 

\subsection{Effect of robot shape on rolling performance}\label{sec:performance}  

\begin{figure*}
\centering
\includegraphics[width=0.7\linewidth]{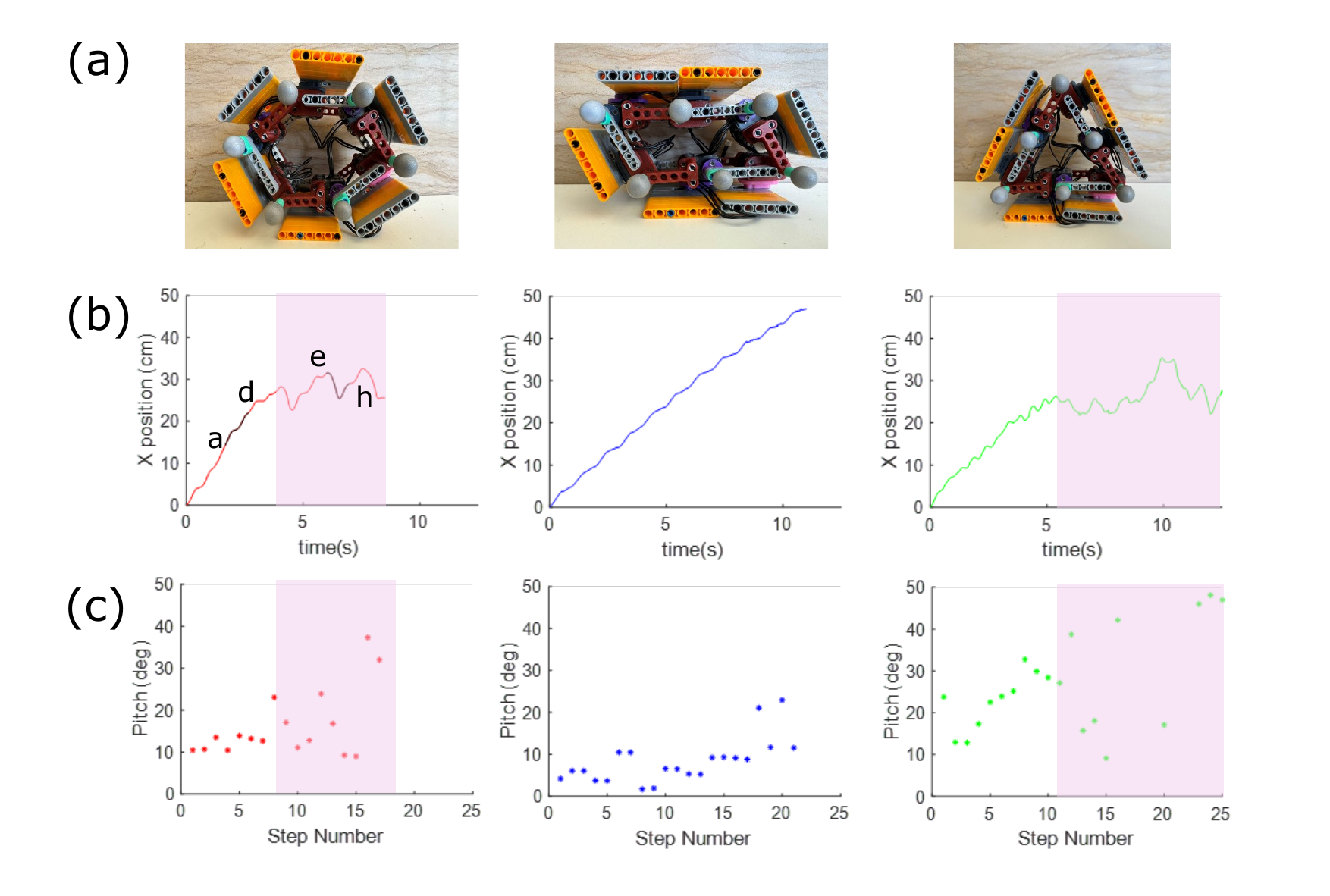}
\caption{Representative trials of the robot locomotion on sand with three different shapes: {\textbf{Left}}: Hexagon; {\textbf{Middle}}: Quadrilateral; {\textbf{Right}}: Triangle. {\textbf{(a)}} images of each shape. {\textbf{(b)}} Experimentally-measured robot $x$-direction position vs. time. The highlighted segment labeled with letter a to d corresponds to experimental images a-d in Fig.5. The highlighted segment labeled with letter e to h corresponds to experimental images e-h in Fig.5. {\textbf{(c)}} Instantaneous body pitch angle measured from each step on sand. In (b) and (c), pink shaded regions represent the ``failure region'', where the robot oscillated in place and not able to effectively move forward. 
} 
\label{Fig.Step-wise-Displacement}
\end{figure*}

Tracked robot speed on rigid ground showed that among the 3 shapes 
Hexagon exhibited the fastest forward speed ($v_x$ = 12.3 $\pm$ 3.7 cm/s), followed by Triangle ($v_x$ = 9.4 $\pm$ 2.2 cm/s), and Quadrilateral being the slowest ($v_x$ = 5.6 $\pm$ 1.6 cm/s). Here $v_x$ is the magnitude of rolling speed along $x$ direction (Fig. \ref{Fig.structure}c), averaged through each step. The recorded robot speeds on rigid ground are shown in Table. {\ref{tab:performance_updated}}. 

Interestingly, the robot's rolling performance on sand was significantly different from its performance on rigid ground. Specifically, the best-performing shape on rigid, Hexagon, failed to move past 20 cm in $83\%$ of the trials, as shown in Table. {\ref{tab:performance_updated}}, with an averaged speed\footnote{Here the averaged robot speed was computed from steps prior to the stopping criteria: (1) completed the entire test range; 2. collided with the side wall of the trackway; 3. fell sideways or exhibited significant turning such that cannot be reliably tracked by the MoCap system; 4. stuck in place for 2 consecutive stride cycles.} of $v_x$ = 1.3 $\pm$ 1.0  cm/s.  Similar to Hexagon, the Triangle shape exhibited a relatively high failure rate (67$\%$) on sand, with an averaged speed of $v_x$ = 1.5 $\pm$ 1.3 cm/s (Table. {\ref{tab:performance_updated}}). On the other hand, the ``worst-performing'' shape on rigid ground, Quadrilateral, was able to successfully travel across the entire length of the testing field without any failure, with an averaged speed of $v_x$ = 3.4 $\pm$ 0.6 cm/s (Table. {\ref{tab:performance_updated}}).

\begin{table}[h]
    \centering
    \caption{Performance comparison of different shapes on rigid ground and sand. Here the failure rate was computed as the percentage of trials that failed to traverse more than 20 cm. 
    } \label{tab:performance_updated}
    \begin{tabular}{|p{1cm}|p{1cm}|p{1cm}|p{1cm}|p{1cm}|p{1cm}|}
        \hline
        \textbf{Shape} & \textbf{Averaged speed on rigid ground (cm/s)} & \textbf{Failure rate on rigid ground (\%)}  & \textbf{Averaged speed on sand (cm/s)} & \textbf{Failure rate on sand (\%)} & \textbf{Traversed distance on sand before failure (cm)} \\
        \hline
        Hexagon & 12.3$\pm$3.7 & 0 & 1.3$\pm$1.0 & 83.3 & 9.5$\pm$9.2 \\
        \hline
        Quadri- lateral  &  5.6$\pm$1.6 & 0 & 3.4$\pm$0.6 & 0 & N/A\\
        \hline
        Triangle &  9.4$\pm$2.2 & 0 & 1.5$\pm$1.3 & 66.6

        & 15.6$\pm$14.0 \\
        \hline
    \end{tabular}    
\end{table}

A close analysis of the robot fore-aft position vs. time showed that the Hexagon and Triangle could effectively move forward initially in all trials. However, after a few steps they would quickly enter the failure mode (Fig. {\ref{Fig.Failure mode}}b, pink shaded region), where the robot oscillated in place and could not roll forward. It was also noticed that when the robot could reliably move forward, its body pitch angle was generally small, but as it entered the oscillating failure mode, its body pitch angle from the previous step became significantly larger (Fig. {\ref{Fig.Failure mode}}c). 

\subsection{Hypothesized failure mechanism for sand-rolling}\label{sec:phenomena}

\begin{figure}[bthp!]
\centering
\includegraphics[width=0.6\linewidth]{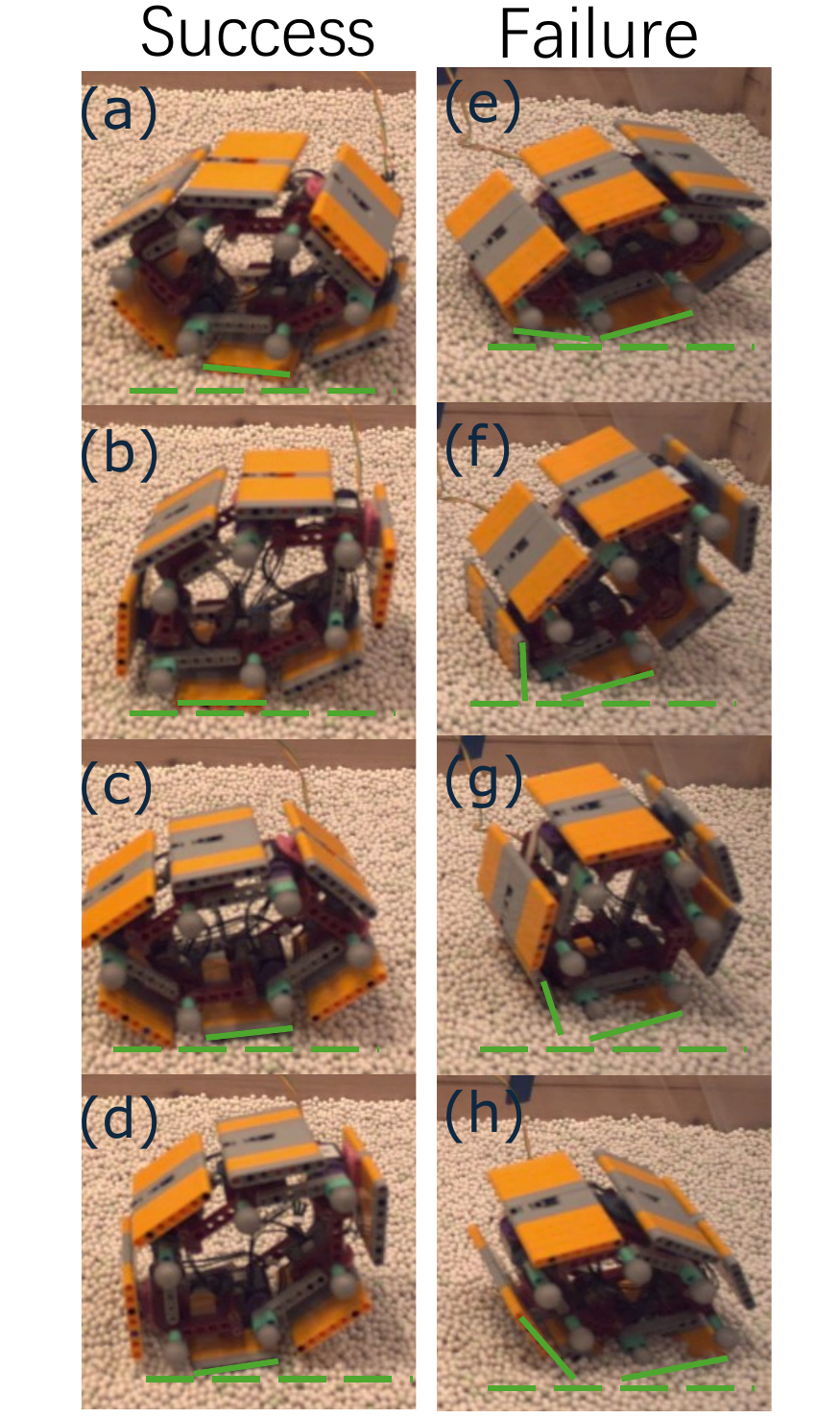}  
\caption{
Image sequence from experimental footage showing a representative trial of Hexagon. \textbf{Left}: (a) to (d) illustrate 2 successful steps (corresponding to Fig. \ref{Fig.Step-wise-Displacement}b highlighted tracking data segment labeled a-d)  where the robot continuously rolled forward. \textbf{Right}: (e) to (h) illustrate 2 failing steps (corresponding to Fig. \ref{Fig.Step-wise-Displacement}b highlighted tracking data segment labeled e-h)  where the robot oscillated back and forth in place. 
}
\label{Fig.Failure mode}
\end{figure}

Intrigued by the high failure rate of the robots  on sand for the Hexagon and Triangle shapes, we examined the interactions between the robot and the deformable substrate, to investigate possible failure mechanisms. Fig. \ref{Fig.Failure mode} shows a representative trial of the Hexagon rolling robot initialling successfully rolling on sand (Fig. \ref{Fig.Failure mode} a-d), but eventually exhibited a failure mode where the robot oscillated in place (Fig. \ref{Fig.Failure mode} e-h).

An interesting phenomena observed from the Hexagon shape experiments was that during the failure steps on sand, the robot body was supported by two sand-contacting segments (Fig. \ref{Fig.Failure mode}e-h, green solid lines) that oscillated towards opposite directions, producing a ``clam-like'' motion and causing the robot to dig into the sand. This was distinct from robot's movement pattern found on rigid surface and in successful traversal steps on sand, where the robot body was mostly supported by a single segment laying relatively flat on the substrate (Fig. \ref{Fig.Failure mode}a-d, green solid lines). 


As a result, the body pitch (defined as the angle measured from horizontal plane to the supporting segment) was significantly higher during the failing steps (Fig. \ref{Fig.Failure mode}e-h) as compared to the successful ones (Fig. \ref{Fig.Failure mode}a-d). For robots with the Hexagonal shape, tracked robot body pitch, $\beta$, measured from the horizontal sand surface to the robot's lowest body segment at supporting segment-switching moments, remained around 10.0$\degree$ $\pm$ 4.1$\degree$ for successful steps (\ie with a stride length 
greater than 2 cm), but increased significantly to 38.2$\degree$ $\pm$ 18.9 $\degree$ for failing steps (\ie with a stride length below 2cm) (Fig. \ref{Fig.Failure mode}c). 

\subsection{Model-predicted critical pitch angle for producing sustainable rolling on sand}\label{sec:mechanism}



Based on the observed high body pitch during failure mode, in this section we propose a simple geometric model to reason about the necessary condition for sustainable rolling, and explain how body pitch influences robot's rolling performance. 

\begin{figure*}[htbp]
\centering
\includegraphics[width=0.65\linewidth]{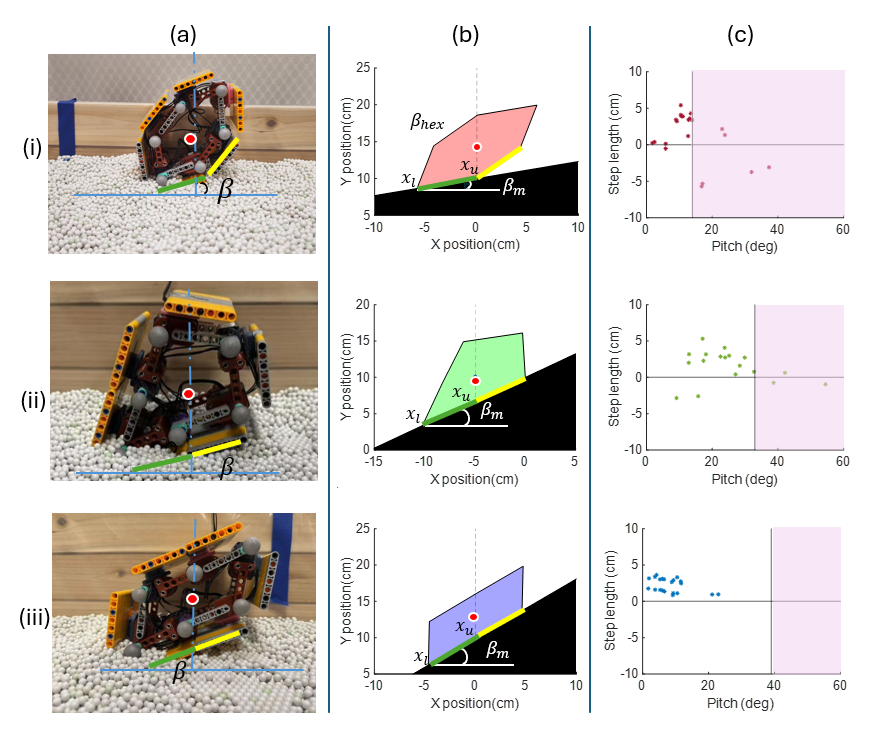}
\caption{Proposed model to explain the failure mechanism for the three shapes: \textbf{(i)}: Hexagon; \textbf{(ii)}: Triangle; \textbf{(iii)}: Quadrilateral. \textbf{(a)} Experimental images taken at the supporting segment switching moment for each shape. 
\textbf{(b)} Diagrams illustrating the proposed model. The theoretical maximum pitch angle for the robot to produce sustained rolling, $\beta_m$, is computed as the largest slope angle which the robot with each shape can climb up. In (a) and (b), red circle represent robot CoM. Green and yellow highlighted segment represent the current and next supporting segments, respectively. [$x_l$, $x_u$] represent the effective region of supporting polygon (ERSP).
\textbf{(c)} Experimentally-measured robot step length, plotted against the robot's instantaneous body pitch angle at the supporting segment switching moment during each step.
Black vertical line represents the model-predicted critical pitch angle, beyond which the robot would fail to achieve supporting segment switching and sustainable rolling. Pink shaded regions represent the model-predicted failure region. }
\label{Fig.failure mechanism}
\end{figure*}

In this highly-simplified model, the robot mass was assumed to primarily concentrated at the geometric center (Fig. \ref{Fig.failure mechanism}a, b, red circle). To produce sustainable rolling, a necessary condition is that the vertical projection of the robot CoM on the horizontal plane needs to be shifted outside the boundary of the supporting segment (Fig. \ref{Fig.failure mechanism}a, b, segment highlighted in green)~\cite{mcghee1968stability, bretl2008ZMP, kalakrishnan2010fast, E_Soft_Robot}. This condition allows the ``supporting segment switching'', \ie the touchdown of a new supporting segment, and the lift-off of the previous supporting segment.

On rigid flat ground, the supporting segment was always horizontal to the ground, resulting in a $0\degree$ body pitch. As a result, all three shape kinematics was able to effectively shift the robot CoM outside the supporting region, achieving supporting segment switching and producing sustained rolling.

On sand, however, the supporting segment was often not horizontal. Granular substrates like sand are yield-stress~\cite{nedderman1992statics} materials, which behave solid-like if the external stress remains below the yield stress, and flow like fluid when the external stress exceeds the yield stress. As such, during robot rolling, the part of the supporting segment that first reached the sand would begin to deform the substrate, while its penetration depth increased. This created an increased pitch, as if the robot was moving on a slope (Fig. \ref{Fig.failure mechanism}b), and increases the effective region of supporting polygon (ERSP). This induced a challenge: a large body pitch created a high virtual slope, making it more difficult for the robot to shift its CoM outside the supporting region and achieve sustained rolling. 

We can use the model to predict the maximum pitch angle for each robot shape that can achieve the supporting segment switching. To do so, the robot was fixed on a virtual ``ramp'' with slope angle $\beta$ (Fig. \ref{Fig.failure mechanism}b), supported by a single segment with full contact with the ramp surface. For each shape, given the robot's kinematics, the model computed the projection of the robot CoM on the ramp surface, and determined the largest ramp angle, $\beta_m$, where the CoM projection can be moved outside the ERSP, [$x_l, x_u$] (Fig. \ref{Fig.failure mechanism}b). Here $x_l$ and $x_u$ represent the range of supporting segment on the ramp surface. This largest ramp angle is the maximal pitch angle for each robot shape to produce sustainable rolling on deformable sand. At the supporting segment switching moment,  upon the lift-off of the current supporting segment, if the robot instantaneous pitch is larger than $\beta_m$, the robot would fail to achieve the supporting segment switching, fell backward, and started to oscillate in place as shown in Fig. \ref{Fig.Failure mode}e-h. We refer to $\beta_m$ as the critical pitch angle for each shape.

\subsection{Experimentally-measured body pitch for each shape agreed well with model prediction}


The model-predicted critical pitch angle for producing sustainable rolling was $\beta_m(hex)=13.6\degree$ for the Hexagon, $\beta_m(tri)=33.4\degree$ for the Triangle, and $\beta_m(quad)=39.5\degree$ for the Quadrilateral. To test our hypothesis about the failure mechanism, we compared the model-predicted maximum pitch angles with experimental measurements. Fig.~\ref{Fig.failure mechanism}c shows the experimentally-measured robot step lengths plotted against the instantaneous robot pitch angles at the supporting segment switching moment. If our hypothesis was correct, when the robot pitch is larger than $\beta_m$, the robot step length would either decrease to 0 or became negative (\ie falling backwards). 

We found that the experimental measurements agreed well with the model-predicted critical pitch angle: most steps with pitch angle smaller than $\beta_m$ (Fig. \ref{Fig.failure mechanism}c, black vertical line) exhibited positive step length, whereas steps with pitch angle larger than $\beta_m$ exhibited 0 or negative step length. We noted that there were two steps in Triangle shape where failures occurred at relatively small pitch (Fig. \ref{Fig.failure mechanism}c-ii, lower left). We believed this is due to the feedback effect where the robot-substrate interaction during the previous failure steps disturbed local substrate profile such as slope or strength, causing a greater disturbances for future steps. This feedback effect was also observed and reported in several recent studies~\cite{li2009sensitive,liu2023adaptation,shrivastava2020material}. 

The model-predicted $\beta_m$ for explained the experimentally-observed sand-rolling performance among different shapes. Based on the model prediction, the maximal pitch angle for Hexagon and Triangle to produce sustainable rolling was significantly smaller as compared to the Quadrilateral. As a result, the Hexagon and Triangle shapes exhibited higher failure rate on sand, whereas the Quadrilateral exhibited the most stable rolling. The understanding of the failure mechanism also provides valuable insights to inform design adaptations to improve sand-rolling performance (Sec. \ref{sec:strategy}).



\section{Model-informed design adaptation for improved sand-rolling performance}\label{sec:strategy}

\begin{figure}[hbtp]
\centering
\includegraphics[scale=0.35]{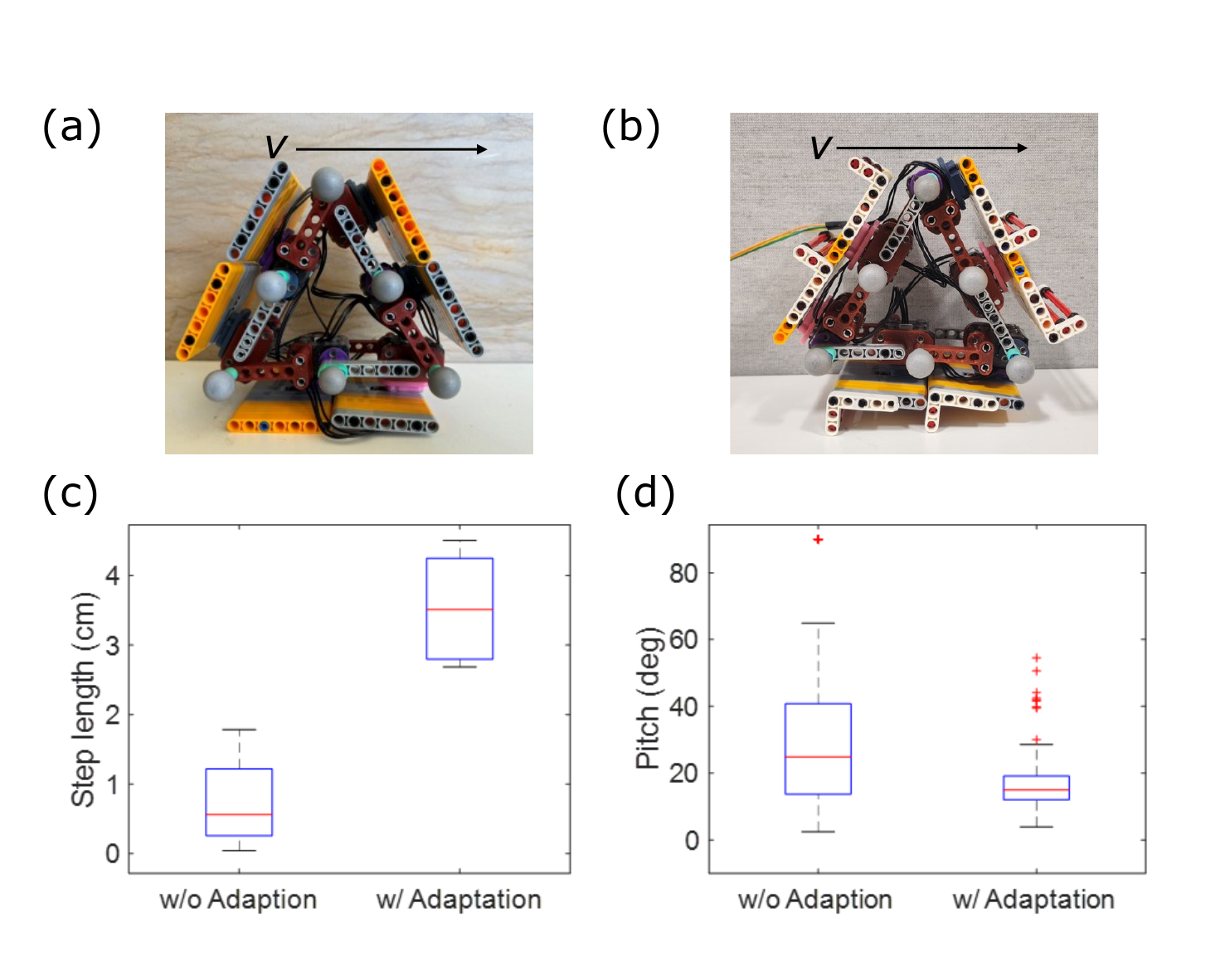}
\caption{The proposed design adaptation to improve robot's sand-rolling performance. \textbf{(a)} A side-view image of the robot without the adaptations.\textbf{(b)} A side-view image of the robot with the proposed adaptation. \textbf{(c)} Comparison of robot's step length on sand between the two different configurations.   \textbf{(d)} Comparison of robot's max pitch angle in each step between the two different configurations.}
\label{Fig.Lbar1}
\end{figure}

Based on the discovered failure mechanism, we proposed a simple design adaptations to improve robot's rolling performance on sand. Specifically, we attached L-shaped brackets connected with horizontal bars on the outside of the segments (Fig. \ref{Fig.Lbar1}b). We expected the added segments to improve the robot's sand-rolling performance through two mechanisms: (i) increasing the sand penetration resistance force~\cite{hill2005scaling, brzinski2013depth} on the supporting segment. The increased penetration resistance force could reduce the sinkage on the back end of the segment, and therefore reducing the associated body pitch to remain below $\beta_m$, preventing the observed oscillation-in-place failure (Fig. \ref{Fig.failure mechanism}). (ii) increasing the sand shear resistive force~\cite{albert1999slow,albert2001stick} on the supporting segment, to reduce the slippage of the supporting segment and increase the robot's displacement during the stance.

To test the effectiveness of the proposed design adaptation, we compared the experimentally-measured robot step lengths and pitch angles between with and without the adaptation for each shape. Results shown that the proposed adaptation could significantly reduce the failure rate for the two ``high speed, high failure rate'' shapes, Hexagon and Triangle, and allowed them to produce sustained rolling on sand. Fig. \ref{Fig.Lbar1} shows the comparison of step length and pitch angle for the Triangle shape on sand. With the simple adaptation, the averaged pitch angle was reduced by 30$\%$, improving the averaged step length by more than 200$\%$. In addition, with the adaptation the failure rate was decreased from 66$\%$ to 0. {\Revision Similarly, the adaptation reduced the averaged pitch angle by over 50$\%$ for Hexagon, mitigating the previously-observed locomotion failures.}  The significantly improved step length and success rate demonstrated that a better of understanding the failure mechanism could enable the robot to utilize extremely simple adaptations to achieve fast and robust rolling on deformable sand.

\section{Model generalization and broader applicability}\label{sec:General Shape}

{\Revision 
In this section, we extend our model analysis to predict rolling outcomes for general shapes beyond the three specific shapes tested in experiments. Fig. \ref{Fig.Generalized model}a-i illustrates a general robot shape at the rolling moment, consisting of 6 equal-length segments, with the opposite sides parallel to each other. This robot shape can be specified using the inner joint angles, $\alpha, \zeta, \gamma \in (0, \pi]$. Since given $\alpha$ and $\zeta$, $\gamma$ can be determined through the geometric constraint, $\gamma=2\pi-\alpha-\zeta$, we define the shape space using only the two adjacent inner angles of the supporting segment, $\alpha$ and $\zeta$ (Fig. \ref{Fig.Generalized model}a-ii).

\begin{figure}[bthp!]
\centering
\includegraphics[width=0.8\linewidth]{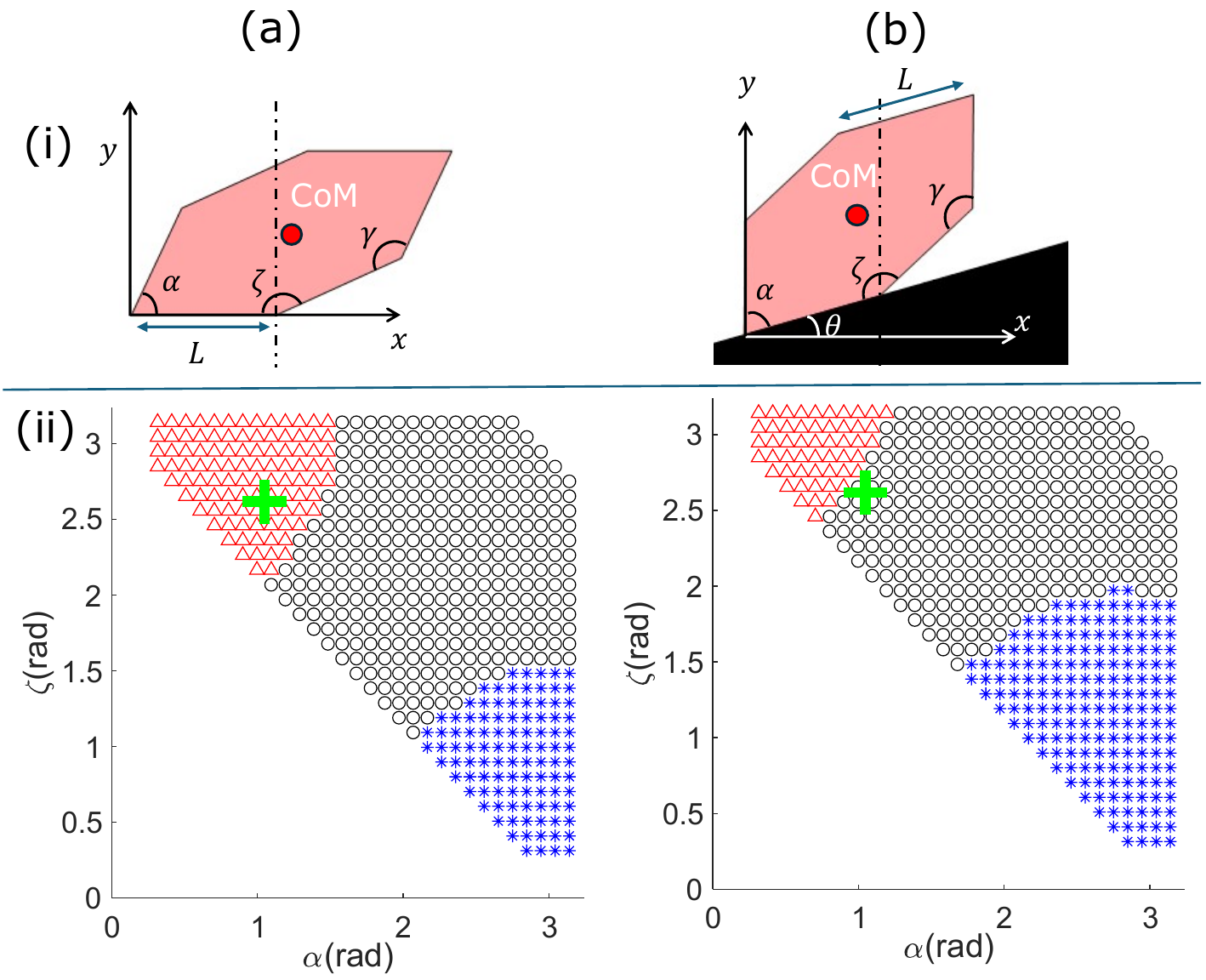}  
\caption{ 
Model-predicted robot rolling performance for generalized shapes on (a) level and (b) sloped surface (20 degree inclination). \textbf{(i)} shows diagrams of the generalized robot rolling shape; \textbf{(ii)} shows model-predicted rolling outcomes for all shapes within the shape space $(\alpha,\zeta)$. Red, blue, black markers represent that the robot can roll towards +x, -x, and cannot roll. Green cross represents the hexagon shape used in experiment, $P(\pi/3, 5\pi/6)$.}
\label{Fig.Generalized model}
\end{figure}

Using the simple geometric model described in Sec. \ref{sec:mechanism}, we can determine whether a specific shape, $P(\alpha,\zeta)$, could achieve the rolling motion. 
For a robot to roll on level surface, it needs to shift its CoM outside the supporting segment region. This translates to $cos(\alpha)-cos(\zeta)>1$ for rolling towards $+x$ direction (Fig. \ref{Fig.Generalized model}a-ii, red markers), and $cos(\alpha)-cos(\zeta)<-1$ for rolling towards $-x$ direction (Fig. \ref{Fig.Generalized model}a-ii, blue markers). Fig. \ref{Fig.Generalized model}ii illustrates the model-predicted rolling outcomes for all shapes in the $(\alpha,\zeta)$ space, on level surfaces (Fig. \ref{Fig.Generalized model}a). 

For a robot to roll on sloped surfaces (Fig. \ref{Fig.Generalized model}b-i), the condition to achieve rolling becomes $cos(\alpha+\theta)-cos(\zeta-\theta)>cos(\theta)$ for uphill, and $cos(\alpha+\theta)-cos(\zeta-\theta)<-cos(\theta)$ for downhill. 
Fig. \ref{Fig.Generalized model}ii illustrates the model-predicted rolling outcomes for all shapes on a sloped surface with inclination $\theta$ (Fig. \ref{Fig.Generalized model}b). We noticed that as the inclination angle, $\theta$, increases, the set of shapes that could roll uphill (Fig. \ref{Fig.Generalized model}b-ii, red markers) decreases, whereas the set of shapes that could roll downhill (Fig. \ref{Fig.Generalized model}b-ii, blue markers) increases. Interestingly, the hexongon shape used in our experiment, $P(\pi/3, 5\pi/6)$, was predicted to be able to achieve rolling on the rigid flat ground (Fig. \ref{Fig.Generalized model}a-ii, green cross), but was no longer able to roll up the 20-degree slope (Fig. \ref{Fig.Generalized model}b-ii, green cross).
This extended analysis, while extremely simple, allows us to begin to reason about how body shapes can influence robot's rolling performance on different terrains. Future work can further extend the model to analyze how additional factors like mass distribution and payload volume capacity can be designed to satisfy locomotion requirements.
}

\section{CONCLUSION}

In this study, we developed a novel sand-rolling robot that is capable of traversing both rigid and deformable surfaces with different body shapes. We investigated how the robot body shape affects the rolling performance on sand, and found that Quadrilateral shape produced the most stable rolling, whereas the Hexagon and Triangle shapes can generate effective rolling initially but exhibited frequent failure after a few steps. Through the analysis of robot dynamics and interaction with the deformable substrates, we discovered that the ability for the robot to produce sustainable rolling on sand was largely dependent on the pitch angle of the robot during supporting segment switching. We developed a simple model that allows prediction of the critical pitch angle for each shape to effectively achieve sustained rolling. The model prediction agreed well with experimental measurements, and informed a simple design adaptation that successfully reduced the pitch angle and failure rate for Triangle and Hexagon, and improved their rolling speeds on sand by more than two folds. 
The insights gained from this study can guide novel design and locomotion control strategies for robots to traverse deformable substrates where traditional robots may struggle. 

\clearpage
\bibliographystyle{./bibliography/IEEEtran.bst}
\bibliography{./bibliography/terradynamics.bib, ./bibliography/mudforce.bib, ./bibliography/mudsensing.bib, bibliography/qian, bibliography/rolling}


\clearpage

\end{document}